# A Foundation Model for Wearable Movement Data in Mental Health Research

Franklin Y. Ruan*, Aiwei Zhang*, Jenny Y. Oh, SouYoung Jin, and Nicholas C. Jacobson

***Abstract*— Wearable movement data is collected by nearly all commercially available smartwatches and is a valuable resource for mental health research, reflecting fine-grained temporal behavioral trends. Despite its promise, the development of foundation models for health wearable modeling remains limited when compared to clinical image and text analysis. We designed transformers with patch embeddings and used self-supervised masked autoencoder pretraining on minute-level week-long actigraphy (physical activity intensity measurement) sequences to develop and evaluate the Pretrained Actigraphy Transformer (PAT). PAT is an open-source foundation model for wearable movement time series that combines week-long temporal modeling, psychiatric outcome evaluation, and reproducibility on public data. Pretrained on data from 21,538 U.S. participants in a nationally representative cohort from the National Health and Nutrition Examination Survey (NHANES), PAT consistently outperformed non-foundation-model baselines across mental health prediction tasks—including benzodiazepine and SSRI use, depression, and sleep abnormalities. During the benzodiazepine medication usage prediction task, PAT demonstrated the largest improvement over non-foundational deep learning models commonly used for time-series modeling (i.e., 55.6% improvement over the LSTM, 21.4% improvement over the 1-D CNN, 14.8% improvement over the ConvLSTM). Beyond predictive accuracy, PAT provides interpretable attention maps highlighting specific periods of daily activity most important for clinical predictions, offering model transparency and potential clinical insights. The results suggest that PAT offers an easy-to-deploy, adaptable and scalable solution to advance clinical insight from wearable sensor data for researchers and clinicians.**

*These authors contributed equally to the paper.

This research was supported by the National Institute of Mental Health (NIMH) and the National Institute of General Medical Sciences (NIGMS) grant R01MH123482.

Franklin Y. Ruan was with Dartmouth College, Hanover, NH 03755 USA. He is now with the Carle Illinois College of Medicine at the University of Illinois Urbana-Champaign, Urbana, IL 61801 USA.

Aiwei Zhang was with Dartmouth College, Hanover, NH 03755 USA. She is now with the School of Interactive Computing at the Georgia Institute of Technology, Atlanta, GA 30332 USA.

Jenny Y. Oh was with Dartmouth College, Hanover, NH 03755 USA. At the time of submission, she is affiliated with Liver Diseases Branch, National Institute of Diabetes and Digestive and Kidney Diseases, National Institutes of Health, Bethesda, MD, USA.

SouYoung Jin is with the Department of Computer Science at Dartmouth College, Hanover, NH 03755 USA.

Nicholas C. Jacobson is with the Center for Technology and Behavioral Health and Departments of Biomedical Data Science and Psychiatry at Geisel School of Medicine, Lebanon, NH 03766 USA, and the Department of Computer Science at Dartmouth College, Hanover, NH 03755 USA.

## I. INTRODUCTION

WEARABLES have become an integral part of everyday life, with commercial smartwatches and fitness trackers [1], [2] continuously collecting streams of physiological and behavioral data from millions of users worldwide [3]–[5]. Movement-based actigraphy—minute-level accelerometer data collected from the wrist or hip—has played an important role in research and clinical applications since the 1970s [6]–[8]. Originally developed for sleep and circadian rhythm assessment in free-living environments, actigraphy is now widely used to study a broad range of health domains, including psychopathology [9], [10], chronobiology [11], [12], and disease risk management [4], [13]–[16]. The unique combination of passive collection, longitudinal coverage, and practicality makes it a powerful tool for capturing behavioral patterns at scale. At minute-level resolution over multi-day windows, actigraphy captures clinically informative phenomena, such as circadian fragmentation, psychomotor slowing, and day-to-day variability in activity rhythms [9], [10], [17], [18], providing sensitive markers of disease progression, treatment response, and risk stratification [4], [15], [19]. Despite its promise, analytical approaches for actigraphy have not kept pace with fields like clinical text and image processing. Classical feature-engineering pipelines relied on handcrafted time- and frequency-domain features, statistics, and rule-based detectors; while transparent, these approaches are task-specific and sensitive to cohort and device differences [20]. Deep learning introduced convolutional and recurrent architectures that learn directly from raw sequences, but convolutional networks have limited receptive fields [21], [22], while recurrent models face gradient degradation and capacity constraints over long windows [23]–[25]. Across both classical and deep learning approaches, generalization beyond training cohorts has been limited, and models commonly assume fixed input lengths or require heavy preprocessing. Furthermore, many models for actigraphy focus on second to minute long sequences for activity recognition [26], [27], leaving multi-day dependencies and circadian structure that emerge only at minute-level resolution over week-long windows largely unaccounted for [21], [28], [29]. These limitations motivate the development of foundation models that can learn generalizable representations from long-range temporal dependencies.

The objective of this study was to develop and evaluate a foundation model specifically designed for wearable movement data. We introduce the Pretrained Actigraphy Transformer (PAT), a foundation model leveraging a trans-former encoder backbone with patch embeddings to efficiently model week-long activity sequences, while employing masked autoencoder-style pretraining to learn robust, generalizable representations from large-scale, unlabeled data. We pretrained PAT on minute-level wearable movement data with over 217 million actigraphy data points from 21,538 participants in a nationally representative sample (NHANES) [30]. We fine-tuned and evaluated PAT across multiple mental health-relevant outcomes. PAT is a fully open-source foundation model, releasing pretrained weights, code, and training data designed for week-long actigraphy in psychiatric and sleep research.

We list the unique contributions of PAT as follows: (1) Open source foundation model for long-sequence actigraphy. (2) Demonstrated generalizability on five clinically diverse mental health-related tasks spanning medication, psychiatric, and sleep domains. (3) Easy-to-deploy with high-end hardware, variable input length, and device-agnostic. (4) Built-in model explainability and transparency.

## II. Related Work

### A. Foundation Models for Wearable Time-Series

The development of foundation models for wearable sensor data has accelerated rapidly. Several recent works are relevant to the positioning of PAT (see Table I).

**Industry-scale proprietary wearable foundation models.** Abbaspourazad et al. [31] trained contrastive foundation models for PPG and ECG from approximately 141,000 Apple Watch participants. Narayanswamy et al. [32] (LSM) extended this to multimodal wearable data, pretraining a masked autoencoder on 26 engineered per-minute features from over 165,000 individuals and up to 40 million hours, establishing scaling laws for wearable foundation models. Erturk et al. [33] developed foundation models for higher-level behavioral signals, evaluating across 57 health tasks, while SensorLM [34] connected wearable sensors to natural language through contrastive and generative pretraining for 60 million hours. These models demonstrate the potential of scale but rely on proprietary datasets and do not release pretrained weights.

**Motion-specific foundation models.** RelCon [26] introduced relative contrastive learning for raw 3-axis accelerometry from approximately 87,376 Apple Heart and Movement Study participants, achieving strong performance on activity recognition and gait regression. Their training code is released but pretrained weights are not, as their dataset is proprietary. Moreover, RelCon is designed for a different modeling scope than PAT; RelCon operates on short segments (2.56 seconds, max input window: $\sim$10 mins) for activity recognition, rather than modeling multi-day behavioral patterns relevant to psychiatric outcomes. UniMTS [27] introduced a contrastive framework aligning physics-simulated accelerometer data with LLM-enriched text descriptions, achieving strong zero-shot activity classification across diverse device placements and orientations. Like RelCon, however, UniMTS operates on short temporal windows ($\sim$7 seconds of simulated skeleton data) and targets discrete physical activity recognition rather than extended behavioral modeling for clinical health outcomes. Other works have examined additional short-segment on-device foundation models [35] and cross-domain transfer from speech foundation models to wearable time series [36].

**Sleep and behavioral foundation models.** Stanford Sleep Bench [37] systematically benchmarked self-supervised pretraining strategies across over 17,000 polysomnography recordings, finding comparable performance across multiple pretraining methods. SleepFM [38] trained a multimodal sleep foundation model on over 585,000 hours of PSG data, predicting over 130 conditions. These studies operate on clinical polysomnography rather than consumer wearable actigraphy. JETS [39] proposed a JEPA-style architecture with a Mamba backbone for behavioral time series from 16,500 individuals, handling irregular sampling through latent-space prediction, though its dataset and pretrained weights are proprietary.

### B. Positioning PAT within the Wearable Foundation Model Landscape

Table I summarizes the key distinctions among recent wearable foundation models. Most existing foundation models either (a) operate on short temporal segments or aggregated features, potentially missing extended behavioral patterns; (b) are trained on proprietary data that external researchers cannot access; (c) do not release pretrained weights; or (d) focus on activity recognition or general health biomarkers rather than psychiatric and sleep-related outcomes.

PAT addresses this gap. It processes 10,080-minute (week-long) actigraphy sequences at minute-level resolution—to our knowledge, the longest temporal context among open wearable foundation models—and is evaluated on psychiatric medication use (benzodiazepines, SSRIs), depression, and sleep abnormalities prediction. PAT is pretrained on publicly available NHANES data and releases all pretrained weights, preprocessing pipelines, tutorials, and code, enabling reproducibility and adaptation by the behavioral health research community.

## III. Methodology

### A. NHANES Dataset and Actigraphy

This study used data from 29,307 participants in the National Health and Nutrition Examination Survey (NHANES) [30], a public longitudinal, nationally representative program of the U.S. population.

NHANES actigraphy data were collected in four survey cycles: 2003-2004 (7,176 participants), 2005-2006 (7,455 participants), 2011-2012 (6,907 participants), and 2013-2014 (7,769 participants), using either hip-mounted uniaxial or wrist-worn triaxial accelerometers. Each record contains at least 10,080 consecutive minutes of activity counts over a 7-day period.

We used data from cohorts 2003-2004, 2005-2006, and 2011-2012 (a total of 21,538 participants) for model pretraining. Then, we used data from 2013-2014 (7,769 participants) for downstream task fine-tuning and analysis.

### B. Data Pre-processing and Task Definition

*1) Data Pre-processing:* During the 2003-2006 cohorts, activity was measured using a hip-mounted ActiGraph AM-7164

TABLE I
COMPARISON OF RECENT WEARABLE AND BIOSIGNAL FOUNDATION MODELS.

| Model | Data Modality | Method | Pretraining Scale | Pretraining Segment* | Max Input Length | Evaluation | Open Wts | Open Data |
|---|---|---|---|---|---|---|---|---|
| Lee et al. (NeurIPS TS4H 2025) | PPG | MAE with U-NET CNN | 47.6K subj., ∼80K hrs | 10 s | 10 s | Cardio and metabolic screening | ✗ | ✗ |
| UniMTS (Zhang et al. NeurIPS 2024) | Motion time series | Contrastive learning | 28.59 hrs $((\sim)14K$ human motion skeletons) | ∼ 7.1 s | 10 s | 18 real-world motion classification benchmarks | ✓ | ✓ |
| Abbaspourazad et al. (ICLR 2024) | PPG, ECG | Contrastive | ∼141K subj., ∼3 yrs | 30/60 s | 30/60 s | General health biomarkers | ✗ | ✗ |
| Narain et al. (NeurIPS TS4H 2025) | Cross-domain (speech → wearable) | HuBERT / wav2vec 2.0 | ∼60K hrs speech | 2/10/60 s | 60 s | Activity, arrhythmia, stress | ✗ | ✗ |
| RelCon (Xu et al., ICLR 2025) | Raw 3-axis accel. | Relative contrastive learning | 87K+ subj., 1B segments | 2.56 s | 10 min[†] | Activity recognition and gait analysis | code only | ✗ |
| LSM (Narayanswamy et al., ICLR 2025) | Multimodal (26 features) | Masked autoencoder | 165K subj., 40M hrs | 5 hrs (300 min) | 5 hrs | Activity and exercise recognition | ✗ | ✗ |
| Sleep Bench (Kjaer et al., 2025) | Multimodal PSG (16 ch.) | Comparing MAE, CL, DAE | ∼13K subj., ∼163K hrs | 300 seconds | ∼8 hrs[†] | Sleep staging, disease prediction | ✓ | ✗ |
| SleepFM (Thapa et al., Nat. Med. 2026) | Multimodal PSG | Leave-one-out contrastive | ∼65K subj., 585K+ hrs | 5 min | 9 hrs | Disease prediction (PheWAS) | code only | partial |
| SensorLM (Zhang et al., NeurIPS 2025) | Multimodal | Contrastive + generative | 103K subj., 59.7M hrs | 1 day (1440 mins) | 1 day | Human activity and general health | ✗ | ✗ |
| Erturk et al. (ICML 2025) | Multimodal (27 behavioral signals) | Regularized contrastive learning | 162K subj., 2.5B hrs | 1 week (168 hrs) | 1 week | 57 health detection tasks | ✗ | ✗ |
| JETS (Xie et al., NeurIPS TS4H 2025) | Single stream (5 domains) | Joint embedding predictive arch. | 16.5K subj., ∼3M person-days | 5K observation counts | 5K observation counts | Diagnostic prediction | code only | ✗ |
| **PAT (ours, 2025)** | **Unimodal actigraphy** | **Masked autoencoder** | **21.5K subj., 217M data pts** | **1 week (10,080 min)** | **variable[‡]** | **5 psychiatric/ sleep tasks** | ✓ | ✓ |

[†]Max input uses aggregation of segment-level embeddings (e.g., attention pooling, sliding window). [‡]Supports variable input lengths via patch-based architecture.

[*]Pretraining Segment denotes the temporal window used for learning objectives

piezoelectric accelerometer, which records uniaxial acceleration as summed activity counts per minute. In 2011-2014, NHANES transitioned to a wrist-worn ActiGraph GT3X+ triaxial accelerometer. To ensure comparability across devices, triaxial measurements were summed into a single count per minute, aligning them with the earlier uniaxial data.

To further increase comparability across devices, we standardized the pretraining datasets from the 2003-2004, 2005-2006, and 2011-2012 cohorts separately using Scikit-learn's StandardScaler. This approach was taken to reduce variations in device-specific measurements, as accelerometer models varied between these years (i.e., hip-mounted or wrist-worn). By transforming participants' raw movement intensities into z-scores, we were able to express each participant's activity relative to all others who used the same device, thereby mitigating the impact of some amplitude discrepancies between different devices. We note, however, that this standardization does not fully eliminate domain differences, and residual device-specific distributional characteristics will remain. As such, the resulting pretraining dataset can be considered heterogeneous, inclusive of device-dependent variation.

We also standardized our downstream task datasets separately, i.e., participants with substantial missing data were excluded according to the NHANES actigraphy quality control guidelines.

In models trained with smoothed data, we used a Savitzky-Golay filter with a window length of 51 time steps and a polynomial order of 3, which has been shown to retain temporal patterns while reducing noise [40].

*2) Downstream Task Definitions:* From the 2013-2014 NHANES cycle, we derived five supervised prediction tasks, summarized in Table II, each framed as a binary classification problem with definitions and clinical significance indicated.

We identified a cohort of 7,769 NHANES 2013–2014 participants with valid actigraphy and medication records. From this pool, we created five binary prediction datasets corresponding to key psychiatric and sleep-related conditions. Labeling criteria were derived from survey responses and validated clinical instruments, such as PHQ-9 for depression and the SLQ-sleep survey for sleep outcomes. Each dataset includes only the subset of participants with complete data for that task. Table II summarizes the label definitions, dataset sizes, and clinical significance of each prediction task.

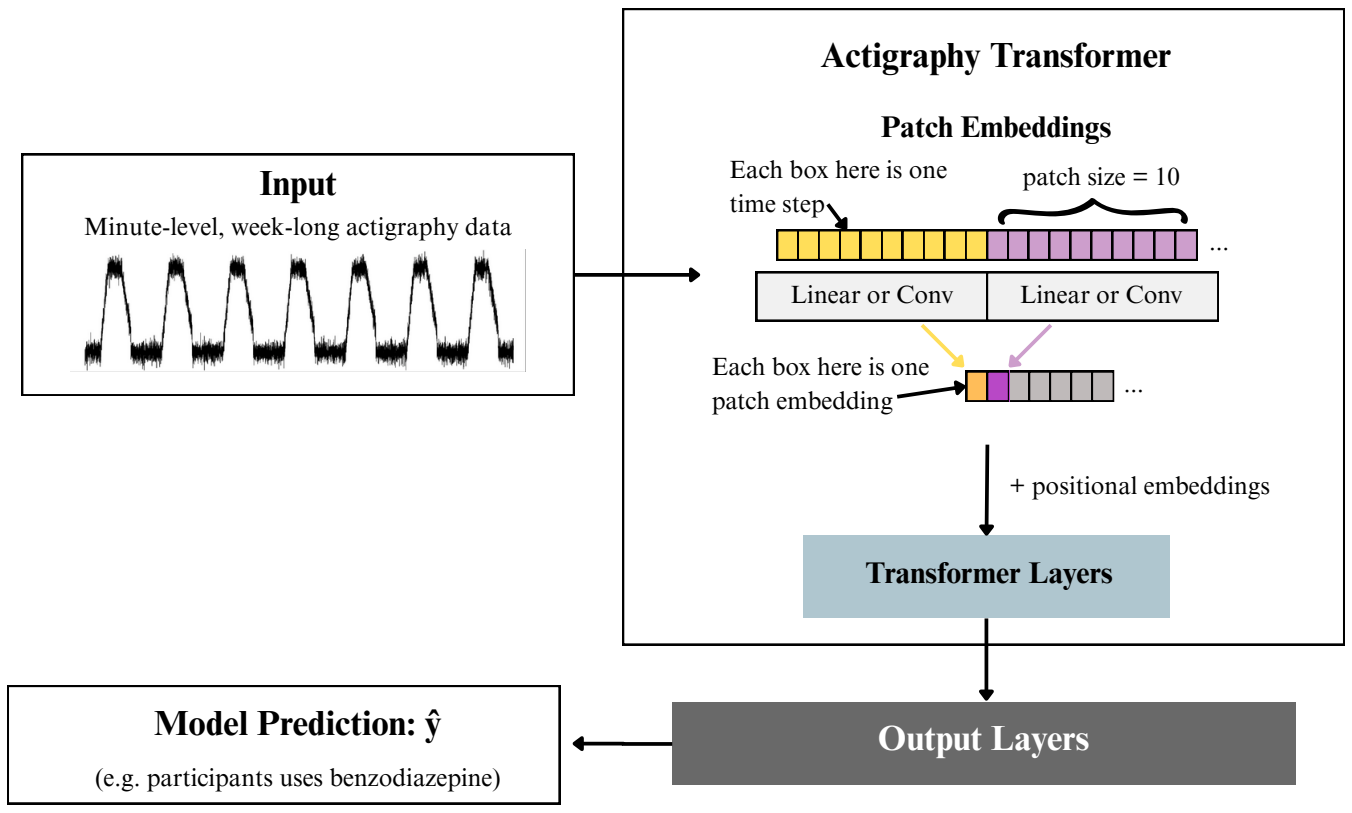

Fig. 1. Model architecture of the Actigraphy Transformer (AT), not pretrained. The figure presents an example case of using an Actigraphy Transformer (AT) for classification. An output layer head is attached to our Actigraphy Transformer and is shown to perform a model classification task.

## C. Modeling Part 1: Designing a Transformer Model for Actigraphy

*1) Actigraphy Transformer Model Design and Patch Embeddings:* Transformers process sequences as a series of input tokens, which translates naturally to the sequential structure of actigraphy data. However, due to the traditional transformer's quadratic complexity at the attention step, it is a non-trivial task to embed 10,000 time-step inputs. Therefore, before pretraining, we first had to design a transformer model architecture capable of effectively processing long minute-level actigraphy data. We refer to this basic unit as the Actigraphy Transformer (AT).

Actigraphy Transformer (AT) draws inspiration from previous work such as the Vision Transformer [42] and PatchTST [43], and applies patching—the sequence is divided into fixed-size, non-overlapping patches, each containing S time steps. The number of resulting tokens is therefore $N = T/S$, where T is the total sequence length. The AT attention layer uses a constant vector size D throughout, and we create patch embeddings by mapping each patch to D dimensions using a trainable linear projection.

*2) Positional Embeddings:* Because transformers are inherently permutation-invariant, we added positional information to each token embedding. We used fixed sine-cosine positional embeddings, which have been shown to perform well in both NLP and computer vision applications [44], [45], to encode the temporal order of patches.

*3) Convolutional Patch Embeddings:* As an alternative to creating patch embeddings through linear projections, we also assessed an architecture that uses 1D convolutional layers to embed each patch. (*Note: PAT models with this embedding have "conv" added to the end of their name.*)

*4) Transformer Encoder:* The embedded patches with positional information are processed by a stack of transformer blocks. Each block contains a multi-headed self-attention layer, a layer normalization layer, a feed-forward network, and a second layer normalization layer.

## D. Modeling Part 2: Masked Auto-encoder Pretraining and Creating Pretrained Actigraphy Transformer (PAT)

*1) Masked autoencoder design:* We employ an asymmetric autoencoder architecture, similar to the masked autoencoder (MAE) pretraining method used for images [45]. In MAE, random portions of the input data are masked, and the model is trained to reconstruct the missing sections.

Consistent with the original MAE approach, masked patches are completely removed before being fed into the encoder (Figure 2). This design significantly increases pretraining speed as we only feed a small subset of the patches to the encoder. Since our encoder will not encounter mask tokens during the fine-tuning step, removing mask tokens during the encoder step has been shown to increase performance. Mask tokens are, however, introduced to the decoder. The input to our decoder, then, is each patch's embedding from our encoder and mask tokens representing the remaining patches.

Fixed positional embeddings are added to all the patches before being inputted into both the encoder and decoder. More information on the specific parameters of the pretraining model can be found in Supplementary Table 1.

*2) Reconstruction objective:* We attempt to reconstruct standardized actigraphy data on the minute level with the decoder, such that the output length is equal to the input at 10,080 time steps. We use a mean squared error at the minute level between the standardized input and the output for our loss function.

*3) Fine-tuning:* After pretraining, the patch embedding layer and transformer encoder layers are extracted and reassembled such (see Figure 2). We refer to this new model as the Pretrained Actigraphy Transformer (PAT). For supervised tasks in this study, we add a task-specific feed-forward classification layer and a sigmoid activation for binary outcomes after PAT. More information on the specific parameters of the fine-tuning pipeline can be found in Supplementary Table 1.

## E. Hyperparameter Tuning

To examine how specific design choices from the pretraining step affect PAT's downstream task performance, we conducted studies on three pretraining parameters: masking ratio during pretraining, input signal smoothing, and the formulation of the reconstruction loss function. These experiments were intended to clarify the contribution of each choice to representation quality and predictive performance in downstream tasks.

All hyperparameter tuning was conducted on a separate training set from the 2013 to 2014 NHANES cohort to prevent data leakage. Specifically, we utilized the benzodiazepine prediction training dataset (see Table II). We created a validation set for hyperparameter tuning by randomly splitting off 1000 participants from the benzodiazepine prediction training data.

For each of the following parameters adjusted during pretraining, we finetuned and evaluated the altered PAT-M on the downstream task of predicting benzodiazepine use from actigraphy.

TABLE II
DOWNSTREAM TASK BINARY PREDICTION, DATASET SIZES, AND CLINICAL SIGNIFICANCE

| Task | Label Definition | # Participants | Clinical Significance |
|---|---|---|---|
| Benzodiazepine use prediction | 1 if participant reported any benzodiazepine medication; 0 otherwise | 7,769 | Benzodiazepines treat anxiety, insomnia, and seizures but carry dependency/misuse risks; may alter sleep–wake rhythm and activity levels detectable in actigraphy. |
| SSRI use prediction | 1 if participant reported taking any selective serotonin reuptake inhibitor (SSRI); 0 otherwise | 7,769 | SSRIs are common antidepressants; they can influence motor activity, circadian timing, and sleep architecture. |
| Depression classification (PHQ-9) | 1 if PHQ-9 score $\geq$ 10; 0 otherwise | 4,800 | PHQ-9 $\geq$ 10 is a validated threshold for major depressive disorder [41]; often linked to reduced daytime activity and altered rest–activity cycles. |
| Sleep disorder classification | 1 if participant reported having a sleep disorder at any time; 0 otherwise | 5,429 | Covers insomnia, apnea, and restless leg syndrome; reflected in disrupted nocturnal and diurnal patterns. |
| Sleep abnormality detection | 1 if participant met any of the following: (1) reported a sleep disorder, (2) regularly slept $\geq$ 12h, or (3) regularly slept $\leq$ 5h; 0 otherwise | 5,429 | Captures diagnosed disorders and extreme sleep durations, each associated with adverse health outcomes and distinct actigraphy patterns. |

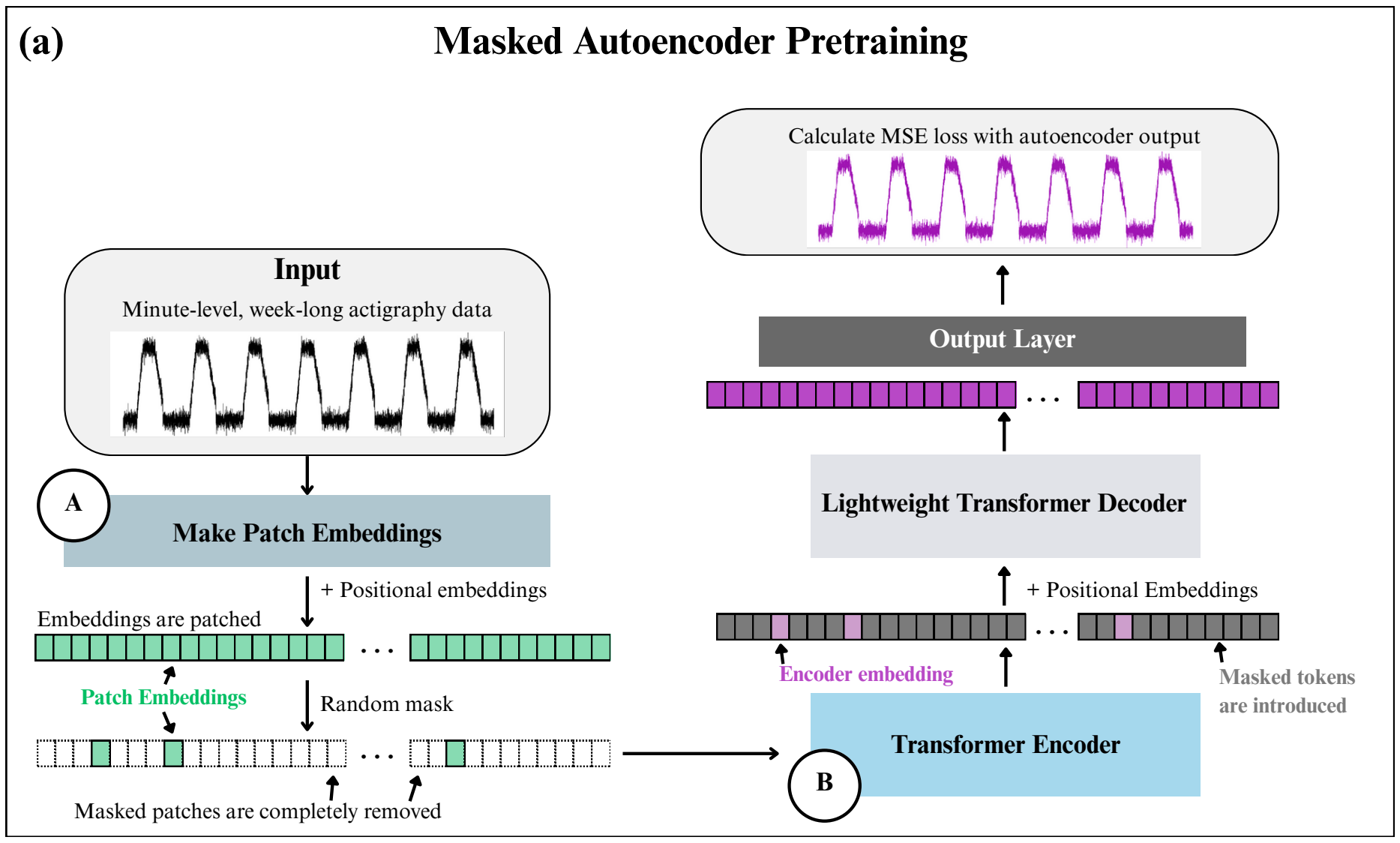

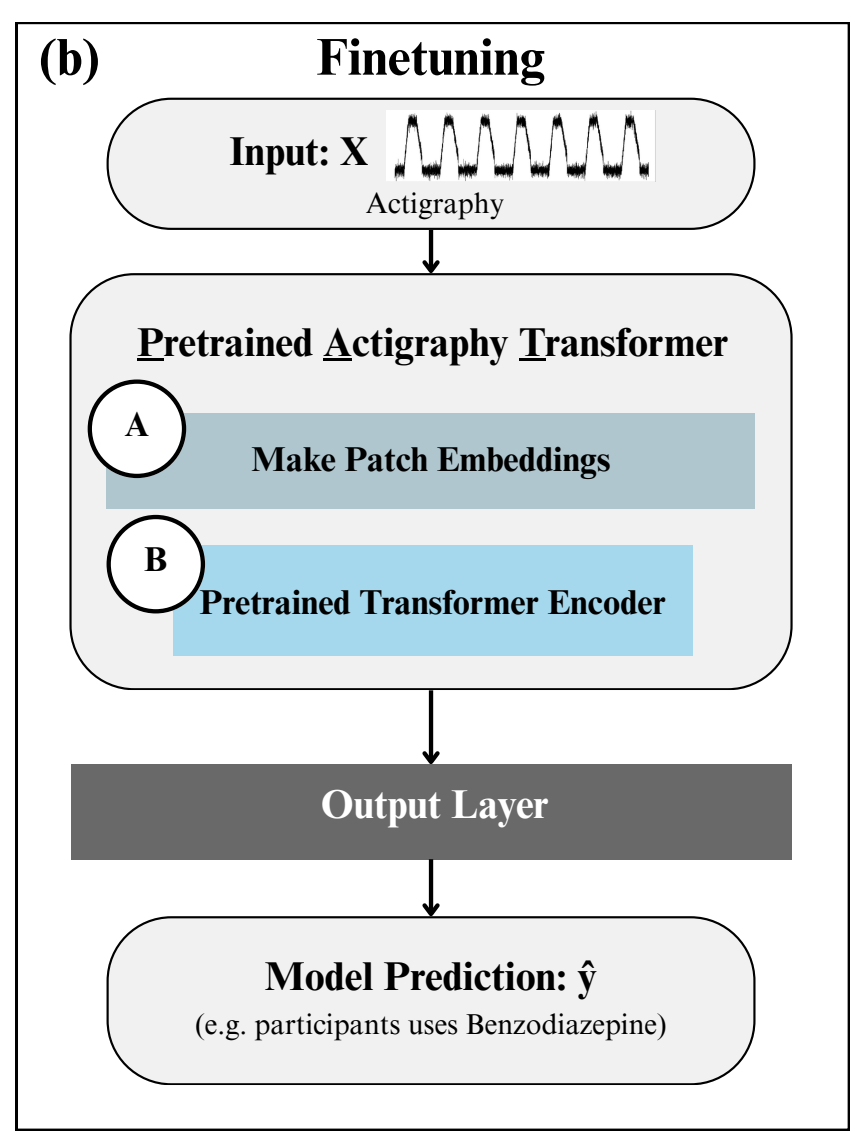

Fig. 2. PAT Pretraining and Fine-tuning. (a) Pretraining: raw actigraphy data is converted into patch embeddings enhanced with positional information. A significant portion of these patches (e.g., 90%) is randomly masked and removed. The remaining patches are fed into a transformer encoder. The encoder's output is then processed by a lightweight decoder to reconstruct the original input data. The pretraining loss is calculated as the mean squared error between reconstructed and original data. (b) Fine-tuning: after pretraining, the patch embedding layer and the pretrained transformer encoder are extracted and assembled as seen in the figure to create PAT. This model can be integrated with various output layers for downstream tasks (e.g., classification).

### *F. Summary of Experimental Design*

*1) Pretraining:* PAT was pretrained using actigraphy data from the 2003–2004, 2005–2006, and 2011–2012 NHANES cohorts, totaling 21,538 participants. Sequences were divided into fixed-size patches, and 90% of patches were randomly masked during training. The encoder processed only unmasked patches, while the decoder reconstructed the full sequence.

*2) Supervised fine-tuning:* All downstream prediction tasks used actigraphy from the 2013–2014 NHANES cohort. For each task, we held out 2,000 participants as a fixed test set. The remaining participants were sampled to create training sets of 500, 1,000, 2,500, and the full available set (denoted as variable n), using stratified sampling to maintain class balance. Twenty percent of each training set was set aside for validation.

*3) Tasks evaluated:* We fine-tuned PAT for five binary classification tasks: benzodiazepine use, SSRI use, depression (PHQ-9 $\geq$ 10), sleep disorder, and sleep abnormality. Task definitions and clinical relevance are summarized in Table II (Section 3.2.2).

*4) Baselines:* We compared PAT against commonly used deep learning models for wearable time series (see Related Work for more details): 1D CNN, LSTM, and ConvLSTM. To further isolate the contribution of large-scale MAE pretraining from the transformer architecture itself, we additionally trained an Actigraphy Transformer (AT) from random initialization without any pretraining, referred to as ScratchTransformer. ScratchTransformer uses an identical architecture to the PAT

TABLE III
EFFECT OF MASKING RATIO ON DOWNSTREAM PREDICTION PERFORMANCE

| Model (Mask Ratio) | Average Score |
|---|---|
| Medium 0.25 | 0.737 |
| Medium 0.50 | 0.707 |
| Medium 0.75 | 0.743 |
| Medium 0.90 | **0.773** |

TABLE IV
IMPACT OF INPUT SMOOTHING ON PAT-M PERFORMANCE

| Model (Smoothing) | Average Score |
|---|---|
| Medium (Smoothed) | 0.741 |
| Medium (Raw) | **0.773** |

encoder — same patch size, embedding dimension, number of attention heads, feedforward dimension and number of transformer layers — with the same classification head, optimizer, learning rate, and training procedure as all other baselines. All baselines used the same input format, preprocessing, and training-validation-test splits as PAT.

*5) Model Interpretability:* We also evaluated the interpretability of PAT using attention heatmaps over time. These visualizations help identify specific periods of the actigraphy sequence that contribute most to each prediction [42].

### G. Evaluation Metric

We use the area under the receiver operating characteristic curve (AUC) as the primary metric for evaluating model performance in binary classification. AUC is threshold-independent and robust to class imbalance. To assess performance across different data availability scenarios, we average each model's AUC score on the held-out test set after fine-tuning on training set sizes of 500, 1,000, 2,500, and N participants. A higher average AUC score indicates better overall performance across dataset sizes.

## IV. RESULTS

### A. Insights from Hyperparameter Tuning

*1) The Optimal Masking Ratio for Actigraphy Input:* We first varied the masking ratio used during pretraining (Table III). Performance improved steadily as the ratio increased, with 90% masking yielding the highest AUC (0.773). This result is consistent with prior work on masked autoencoders, where high masking ratios encourage the encoder to learn broader contextual dependencies.

*2) Smoothing:* Next, we compared PAT models trained on smoothed versus raw actigraphy signals (Table IV). Models using raw data achieved higher AUC (0.773) than those trained on smoothed inputs (0.741).

*3) Loss Function:* Finally, we tested whether the reconstruction loss should be computed only on masked patches or across all patches (Table V). Models trained with mean squared error (MSE) on all patches achieved significantly higher performance (0.773) compared to those trained with MSE only on masked patches (0.541).

TABLE V
EFFECT OF RECONSTRUCTION LOSS FUNCTION ON DOWNSTREAM PERFORMANCE

| Model (Size, Loss) | Avg Score | n=500 | n=1000 | n=2500 | n=4769 |
|---|---|---|---|---|---|
| Medium (MSE-MASK) | 0.541 | 0.437 | 0.515 | 0.560 | 0.652 |
| Medium (MSE-ALL) | **0.773** | **0.753** | **0.764** | **0.786** | **0.788** |

### B. Performance of PAT Across Health-Related Downstream Tasks

We present the results of fine-tuning PAT across five clinically relevant prediction tasks: benzodiazepine use, SSRI use, sleep disorder, sleep abnormality, and depression (Table II).

Performance is reported using area under the ROC curve (AUC), with comparisons made against common models used for time series analysis. Table VI shows detailed results for benzodiazepine usage; Table VII summarizes performance across the remaining four tasks. In these tables, an underlined baseline model suggests that it is the best-performing baseline at each column, and a bolded PAT model suggests that it is the best-performing PAT model at each column.

*1) PAT's Performance on Benzodiazepine Usage Prediction:* Table VI summarizes model performance on predicting benzodiazepine usage from actigraphy data. Baseline model ScratchTransformer-L (PAT-L without any pretraining) achieved an AUC of 0.650. Across all training set sizes, PAT models consistently outperformed baselines, including LSTM, CNN, ConvLSTM, wavelet-based classifiers, and ScratchTransformers. The best-performing model, PAT-L, achieved an average AUC of 0.767, showing a clear advantage over all other models, even when trained on limited data (e.g., n=500).

*2) Multi-Task Summary:* To evaluate PAT's generalizability, we tested its performance across four additional clinically relevant tasks: SSRI usage, history of sleep disorder, sleep abnormality detection, and depression classification. Table VII presents the aggregate AUC scores for each model on these tasks averaged across all training set sizes.

Across all tasks, PAT models outperformed baseline methods, including CNNs, LSTMs, ConvLSTMs, 3D CNNs, wavelet-based classifiers, and ScratchTransformer.

### C. Adjusting for Potential Confounders

Multivariable logistic regression models incorporating PAT model outputs were used to evaluate independent associations with each outcome, adjusting for demographic covariates (age, sex, BMI) and, where applicable, depression severity (PHQ-9). Full results are provided in Supplementary Table 2.

Model predictions were independently associated with benzodiazepine use ($\beta = 0.59$, $p = 0.005$), SSRI use ($\beta = 0.32$, $p = 0.016$), and depression ($\beta = 0.41$, $p < 0.001$). PHQ-9 score and sex were consistent predictors of medication use, while age was an additional significant predictor of SSRI use.

TABLE VI
MODEL PERFORMANCE IN PREDICTING BENZODIAZEPINE USAGE

| Model | n=500 | n=1,000 | n=2,500 | n=5,769 | Avg AUC | Params |
|---|---|---|---|---|---|---|
| LSTM | 0.501 | 0.487 | 0.474 | 0.512 | 0.493 | 15K |
| LSTM (Smoothed) | 0.506 | 0.508 | 0.482 | 0.499 | 0.499 | 15K |
| Wavelet Transform | 0.674 | 0.625 | 0.598 | 0.583 | 0.620 | 10K |
| CNN-1D | 0.621 | 0.630 | 0.640 | 0.637 | 0.632 | 10K |
| CNN-1D (Smoothed) | 0.633 | 0.634 | 0.644 | 0.646 | 0.639 | 10K |
| ScratchTransformer-L | 0.681 | 0.661 | 0.625 | 0.632 | 0.650 | 1.99M |
| ScratchTransformer-S | 0.675 | 0.665 | 0.634 | 0.633 | 0.652 | 287K |
| ScratchTransformer-M | 0.677 | 0.655 | 0.644 | 0.644 | 0.655 | 1.00M |
| ConvLSTM | 0.663 | 0.681 | 0.650 | 0.677 | 0.668 | 1.76M |
| ConvLSTM (Smoothed) | 0.666 | 0.680 | 0.653 | 0.671 | 0.667 | 1.76M |
| CNN-3D | 0.683 | 0.693 | 0.693 | 0.703 | 0.693 | 790K |
| CNN-3D (Smoothed) | 0.677 | 0.695 | 0.696 | 0.719 | 0.697 | 790K |
| PAT-S | 0.706 | 0.718 | 0.677 | 0.703 | 0.701 | 287K |
| PAT Conv-S | 0.737 | 0.711 | 0.722 | 0.735 | 0.726 | 287K |
| PAT-M | 0.743 | 0.745 | 0.742 | 0.745 | 0.744 | 1.00M |
| PAT Conv-M | 0.753 | 0.756 | **0.760** | **0.773** | 0.761 | 1.00M |
| PAT Conv-L | 0.763 | 0.756 | 0.754 | **0.773** | 0.762 | 1.99M |
| **PAT-L** | **0.771** | **0.765** | **0.760** | 0.771 | **0.767** | 1.99M |

TABLE VII
MODEL PERFORMANCE (AVG AUC) ACROSS ACTIGRAPHY TASKS (SSRI, SLEEP DISORDER, SLEEP ABNORMALITIES, DEPRESSION)

| Model | SSRI | Sleep Dis. | Sleep Abn. | Depression | Params |
|---|---|---|---|---|---|
| LSTM | 0.527 | 0.494 | 0.513 | 0.489 | 15K |
| LSTM (Smoothed) | 0.523 | 0.506 | 0.515 | 0.506 | 15K |
| ScratchTransformer-L | 0.523 | 0.508 | 0.492 | 0.536 | 1.99M |
| ScratchTransformer-S | 0.554 | 0.518 | 0.498 | 0.507 | 287K |
| Wavelet Transform | 0.674 | 0.529 | 0.525 | 0.523 | 10K |
| CNN-1D | 0.616 | 0.563 | 0.534 | 0.522 | 10K |
| CNN-1D (Smoothed) | 0.611 | 0.558 | 0.519 | 0.517 | 10K |
| ConvLSTM (Smoothed) | 0.655 | 0.609 | 0.579 | 0.547 | 1.76M |
| ConvLSTM | 0.606 | 0.606 | 0.585 | 0.550 | 1.76M |
| ScratchTransformer-M | 0.669 | 0.571 | 0.543 | 0.536 | 1.00M |
| CNN-3D | 0.677 | 0.608 | 0.606 | 0.583 | 790K |
| CNN-3D (Smoothed) | 0.680 | 0.605 | 0.615 | 0.586 | 790K |
| PAT-S | 0.641 | 0.587 | 0.555 | 0.560 | 287K |
| PAT Conv-S | 0.656 | 0.616 | 0.573 | 0.587 | 287K |
| PAT-M | 0.690 | **0.641** | 0.641 | 0.559 | 1.00M |
| PAT Conv-M | 0.668 | 0.616 | 0.627 | 0.594 | 1.00M |
| PAT Conv-L | 0.695 | 0.631 | 0.659 | **0.610** | 1.99M |
| PAT-L | **0.700** | 0.632 | **0.665** | 0.589 | 1.99M |

For sleep abnormalities, model predictions remained significantly associated with the outcome ($\beta = 0.44$, $p < 0.001$), alongside PHQ-9 score and BMI.

Only for history of sleep disorders were model predictions not significantly associated after adjustment ($\beta = 0.16$, $p = 0.10$), with the outcome instead primarily associated with PHQ-9 score, BMI, and sex.

Overall, PAT model predictions demonstrated independent associations with multiple outcomes after adjustment, including medication use, depression, and sleep abnormalities, but not with sleep disorder.

### *D. Model Explainability*

To improve interpretability, we extracted attention weights from PAT's encoder to visualize which temporal segments most influenced predictions. High-attention regions (red) indicate that the model placed greater importance on those timepoints, whereas low-attention regions (blue) correspond to potentially less informative regions.

*1) Individual Participant Analysis Example:* As seen in Figure 3(b), PAT assigned high attention to extended early-morning inactivity and a delayed wake-up near midday. Midday and evening activity, which appears similar across most participants, received low attention. Figure 4 shows attention patterns for a participant not taking benzodiazepines. Here, the model highlighted sharp and consistent early-morning physical activity as key features. In contrast, low attention was assigned to stable midday and evening activity.

*2) Population Level Analysis:* To extend interpretability beyond single-case visualization, we quantified attention organization across the full benzodiazepine usage prediction task test set. Figure 5 shows PAT attention at the population level organized around recurrent daily structure. At the population level, the mean attention matrix exhibits repeated off-diagonal bands at approximately 24-hour offsets, the mean 1D attention profile shows consistent day-to-day peaks, and the 24-hour overlay indicates that the strongest attention is concentrated around the morning activity transition alongside some smaller secondary patterns. To verify these findings quantitatively, we calculated temporal regularity by measuring autocorrelation of the 1D attention profile at a 24-hour lag, which was consistently positive across participants (mean = 0.36, $p < 0.0001$). We additionally calculated each participant's wake time heuristically from the 7-day actigraphy profile as the time of maximal morning increase in activity between 03:00 and 12:00, and attention within a $\pm$60-minute wake window was compared with attention outside that window. Attention was significantly enriched around inferred wake transitions (mean wake attention ratio = 1.29, Wilcoxon signed-rank test, $p < 0.0001$). Together, these findings suggest that PAT attention is organized around recurrent daily behavioral rhythms rather than being distributed arbitrarily across the week.

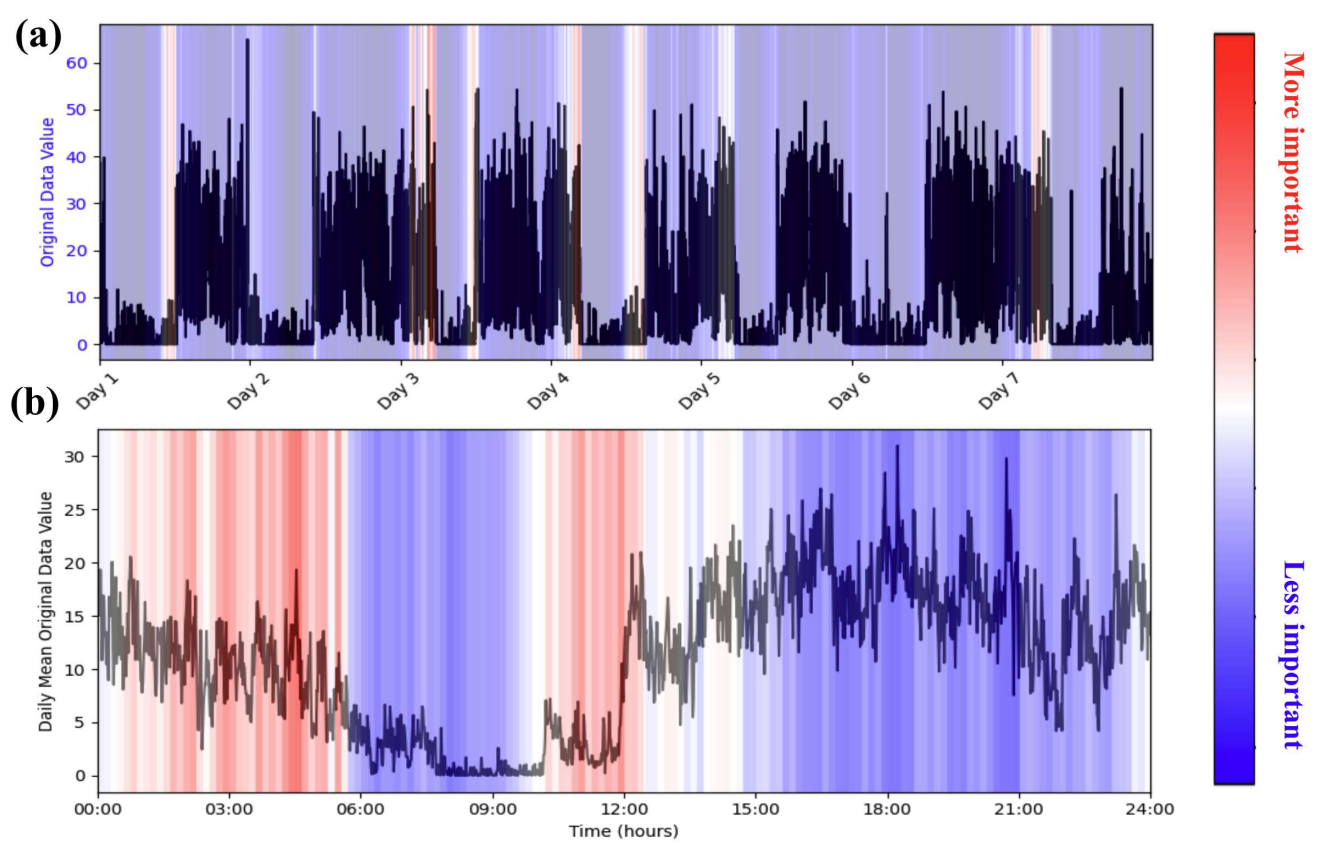

Fig. 3. Example of PAT model explainability for a benzodiazepine user. (a) Week-long actigraphy trace with attention overlay. (b) Daily mean profile with attention overlay. Red regions indicate higher attention and blue regions lower attention.

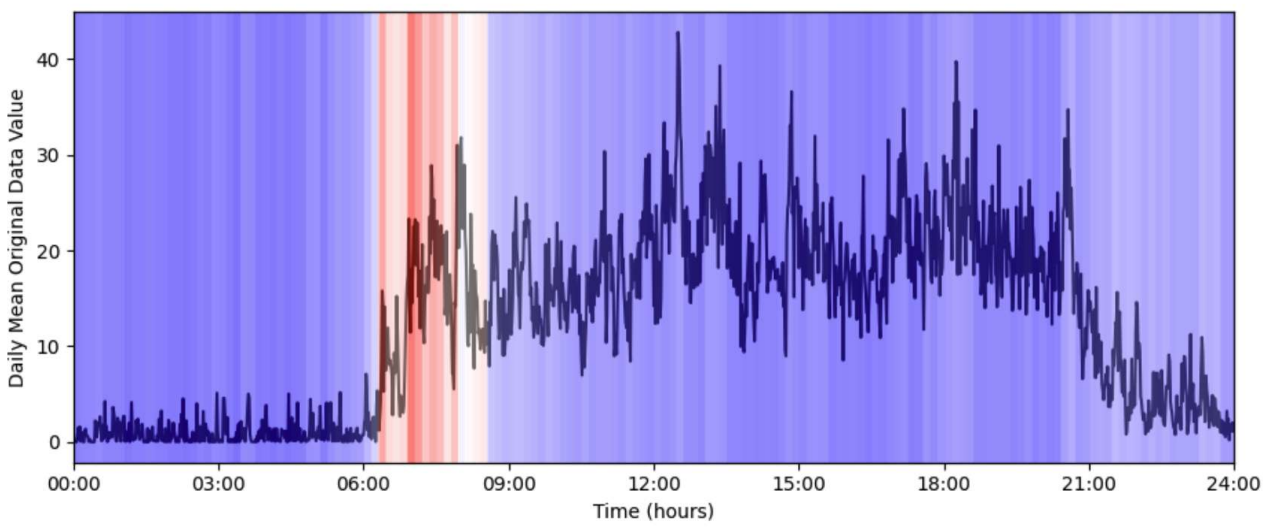

Fig. 4. Example of PAT model explainability for a non-benzodiazepine participant. Daily mean profile actigraphy trace with attention overlay. Red regions indicate higher attention and blue regions lower attention.

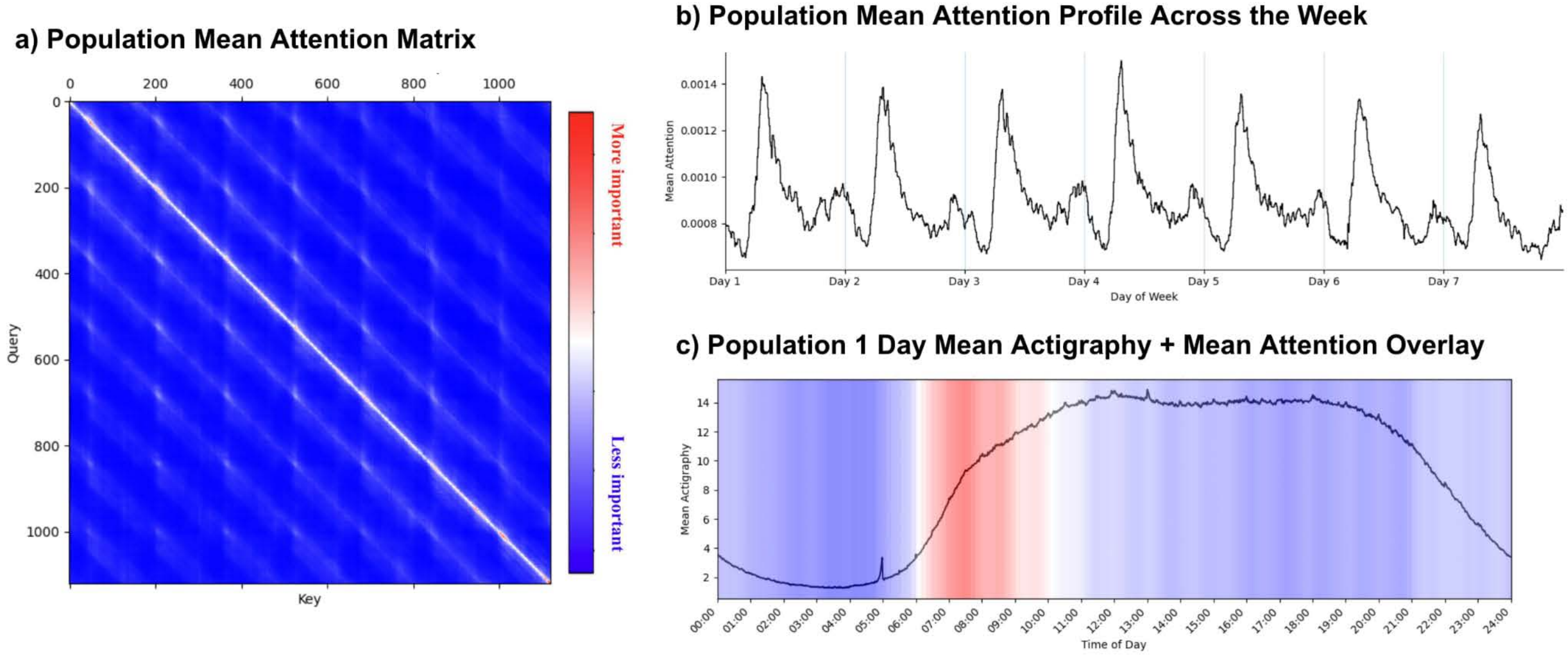

Fig. 5. Population-level attention organization in PAT-L (benzos) across week-long actigraphy. Attention was analyzed from the benzodiazepine use prediction model on the test set. (a) Mean attention matrix across all test participants, showing strong local structure along the main diagonal and repeated off-diagonal bands at approximately 24-hour offsets, consistent with recurrent daily temporal organization. (b) Population mean 1D attention profile across the week, demonstrating repeated daily peaks in attention at similar times across consecutive days. (c) Population mean 24-hour actigraphy profile averaged across the week with mean attention overlay, showing that elevated attention is concentrated around the morning activity transition and broader daytime structure.

*3) Interpreting Ablation:* By using attention weights, we qualitatively examined the effects of the smoothing ablation; we compared attention overlays from PAT-M trained on smoothed versus raw actigraphy input (Supplementary Figure 1). Both models captured broad daily activity structure, indicating that coarse behavioral trends were preserved with smoothing. In comparison, the raw-input model exhibited broader and more distributed attention patterns, consistent with the possibility that unsmoothed actigraphy preserves finer-grained temporal information that may be useful for prediction.

## V. Discussion

In this study, we propose the Pretrained Actigraphy Transformer (PAT), which to our knowledge is the first open-source foundation model to combine week-long, minute-level actigraphy modeling with explicit evaluation on mental health outcomes, while releasing all pretrained weights and training data for reproducibility (see Table I for summative comparison). As transformers are flexible and have few inductive priors, leveraging their potential for complex sequential data allowed PAT to achieve effective embeddings. Together with large-scale pretraining and open-source availability, PAT represents a step toward generalizable and easily deployable human behavior modeling from wearable activity monitors.

### A. PAT as an Adaptable and Accessible Open-Source Model

A core motivation for PAT's development was to create a high-quality, accessible tool to enable collaboration and validation in the behavioral health field.

**Strong Performance in Lower-Data Conditions.** By pretraining on a large, nationally representative cohort (NHANES), PAT provides a powerful starting point for researchers with limited access to large-scale labeled datasets. The release of pretrained models enables rapid fine-tuning for new clinical tasks, offering a key advantage in behavioral health domains where labeled data is often scarce or costly to collect. This approach leverages the strength of foundation models in lower-data conditions (n$\leq$1k). Across all five mental health prediction tasks, PAT variants consistently outperformed baseline models, particularly in lower-data settings.

**Other Considerations Beyond Predictive Performance.** In digital health research, model utility depends not only on discrimination metrics such as AUC, but also on feasibility, transparency, privacy, and robustness to real-world study constraints. PAT was designed to be practical for health researchers with limited computational resources, as all model variants can be fine-tuned on a single GPU, including free ones available on common cloud-based research environments.

To empirically characterize PAT's computational requirements, we benchmarked all models on an NVIDIA T4 GPU (16 GB), a consumer-grade accelerator freely available via Google Colab (Supplementary Table 3). PAT-M (1.00M parameters) completes inference of a 10,080-step sequence in 34.64 ms, which is much faster than recurrent models (LSTM: 117.18 ms; ConvLSTM: 158.85 ms), though slower than purely convolutional models (1D CNN: 8.47 ms; 3D CNN: 7.15 ms). These results demonstrate that PAT achieves a favorable balance between predictive performance and computational speed for practical deployment on accessible hardware.

PAT also offers practical advantages beyond performance alone. The model includes built-in interpretability through attention weight extraction, providing a direct mechanism for examining which temporal regions contributed most to a prediction without requiring separate post hoc explanation tools such as SHAP [46] or LIME [47]. In addition, because PAT can be run and fine-tuned locally, sensitive wearable data do not need to leave the institution where they were collected, which may simplify privacy-preserving workflows. Together, these properties make PAT not only a predictive model, but also a practical foundation for real-world behavioral and clinical research.

**Variable Input Length.** The seven-day input length used in this study was determined primarily by the structure of the NHANES actigraphy dataset, which provides one week of minute-level recordings per participant. This should not be interpreted as an inherent architectural constraint, as PAT can accommodate variable input lengths at the fine-tuning stage or even further downstream. Due to the nature of patch embeddings being non-overlapping and being processed as an independent token with positional embeddings, input length is not restricted to what the model was trained on. The only constraint is that the input length must be evenly divisible by the patch length, which can be resolved by minimal zero padding until the sequence length is evenly divisible. An example of variable-length input with PAT is provided in the fine-tuning demo notebook available in the GitHub repository linked in the Code Availability statement.

### B. Analysis of Confounders, Sleep Disorders, and Sleep Abnormalities

After adjustment for demographic covariates and depression severity using multivariable logistic regression, sleep disorder classification was the only outcome variable not independently associated with PAT-L model outputs ($\beta = 0.16$, $p = 0.10$). In contrast, sleep abnormalities classification remained significantly associated with model outputs ($\beta = 0.44$, $p < 0.001$), indicating preserved independent predictive value (Supplementary Table 2).

Several factors may explain this discrepancy. The sleep abnormalities outcome variable was partly created through the patient's self report on *current* sleep hours, while the sleep disorder outcome variable was positive if participants reported having a sleep disorder at any time in their life. Participants in the sleep disorder class may not be currently experiencing the disorder (e.g., insomnia). Furthermore, formally diagnosed sleep disorders are often actively treated (e.g., continuous positive airway pressure [CPAP] for obstructive sleep apnea), which may attenuate the signal from actigraphy at the time of assessment. As such, for the sleep disorder classification task, the model may be using actigraphy to predict underlying risk factors (confounders), such as elevated BMI for obstructive sleep apnea or depressive symptomatology for insomnia and hypersomnia, rather than the diagnosed condition.

The discrepancy in outcome between these two closely related classes suggests that the choice of the outcome variable requires careful consideration when using associative models such as PAT for clinical phenotyping.

### C. Limitations and Future Work

While PAT demonstrates strong generalization across multiple behavioral health tasks, some limitations warrant discussion. First, the clinical outcome labels used for supervised tasks—such as medication usage or sleep disorder diagnosis—were derived from cross-sectional self-report, questionnaires, and medication-use fields in NHANES rather than gold-standard clinical diagnoses. As such, label noise and subjective bias at the evaluation step may have weakened the observed signal. This limitation primarily affects downstream fine-tuning rather than PAT pretraining, and performance may improve when PAT is applied to datasets with more rigorous outcome labels. More broadly, however, the modest prediction performance of some outcome variables likely also suggests that actigraphy alone provides only partial signal for complex psychopathological features. Indeed, instead of as an individual diagnostic tool, actigraphy may be better used as a complementary behavioral signal in multimodal models or for studying aggregate activity patterns in large cohorts. Furthermore, cross-sectional designs inherently contain confounders and cannot establish causal relationships. Future research should consider cohort studies and randomized controlled trials to better characterize the temporal and causal relationships between actigraphy data and health outcomes. Moreover, the inclusion of both hip-mounted and wrist-worn accelerometers in PAT's pretraining data introduces heterogeneous, device-specific data distributions. While standardization reduces large-scale discrepancies, residual device-specific variation persists and could have introduced confusion in the latent space. However, it is also possible that multi-domain data improved robustness and performance, as found in some studies with self-supervised learning [48], [49]. A promising direction for future work is direct deployment and evaluation of PAT on smartwatch or other edge-device hardware to quantify practical feasibility of device deployment, including inference latency and power consumption in real-world settings.

## VI. Conclusion

We introduced the Pretrained Actigraphy Transformer (PAT), an easy-to-deploy, open-source foundation model tailored for week-long, minute-level actigraphy data. By combining large-scale masked pretraining with flexible transformer architecture, PAT effectively generalizes across a range of clinically relevant behavioral health tasks, including medication usage, sleep disorders, and depression. Its strong performance, interpretability, and deployment-ready design offer a practical step toward scalable digital phenotyping in both research and healthcare settings. Future work may build on PAT to develop multimodal, real-time, and patient-centered models that expand the role of wearable sensing in behavioral health.

## VII. Code Availability

The code for this study is publicly available and accessible via this link (https://github.com/njacobsonlab/Pretrained-Actigraphy-Transformer/). The repository includes links to tutorial/demo notebooks for fine-tuning PAT, model explainability, and masked autoencoder pretraining. The pretrained models themselves are available for download and use.

## VIII. Acknowledgement

This research was supported by the National Institute of Mental Health (NIMH) and the National Institute of General Medical Sciences (NIGMS) grant R01MH123482.

## Conflict of Interest

Nicholas C. Jacobson has received a grant from Boehringer-Ingelheim. Nicholas C. Jacobson has edited a book through Academic Press and receives book royalties, and Nicholas C. Jacobson also receives speaking fees related to his research.

# Supplementary Material

### Supplementary Table 1. PAT architecture and hyperparameters for pretraining and finetuning.

| | PAT-S | PAT-M | PAT-L |
|---|---|---|---|
| ***Input & Patching (shared)*** | | | |
| Input length | 10,080 min | 10,080 min | 10,080 min |
| Patch size (p) | 18 | 18 | 9 |
| Num. patches | 560 | 560 | 1,120 |
| ***Embedding (shared)*** | | | |
| Patch projection | Dense→d | Dense→d | Dense→d |
| Embedding dim (d) | 96 | 96 | 96 |
| Positional enc. | Sinusoidal | Sinusoidal | Sinusoidal |
| ***Encoder (shared)*** | | | |
| Encoder layers | 1 | 2 | 4 |
| Attention heads | 6 | 12 | 12 |
| FFN dimension | 256 | 256 | 256 |
| Dropout | 0.1 | 0.1 | 0.1 |
| ***Pretraining only*** | | | |
| Masking ratio | 90% | 90% | 90% |
| Mask token | Learned | Learned | Learned |
| Decoder layers | 1 | 1 | 1 |
| Decoder heads | 6 | 12 | 12 |
| Decoder FFN | 256 | 256 | 256 |
| Output proj. | Dense(p)+ tanh | Dense(p)+ tanh | Dense(p)+ tanh |
| Loss | MSE | MSE | MSE |
| ***Finetuning only (classification head)*** | | | |
| Pooling | GAP1D | GAP1D | GAP1D |
| Hidden layer | Dense(128)+ ReLU | Dense(128)+ ReLU | Dense(128)+ ReLU |
| Dropout | 0.1 | 0.1 | 0.1 |
| Output | Dense+sigmoid | Dense+sigmoid | Dense+sigmoid |

*Architecture settings for PAT across all three model sizes (PAT-S, PAT-M, PAT-L). Stages shown as "shared" apply to both pretraining and finetuning. The pretraining pipeline adds random masking, a lightweight decoder, and MSE reconstruction loss. The finetuning pipeline replaces the decoder with a classification head (global average pooling → dense(128, ReLU) → dropout → dense(1, sigmoid)). Encoder weights during finetuning are initialized from the corresponding pretrained model. Variable input lengths are supported across all sizes.*

### Supplementary Table 2. Standardized beta coefficient estimates for potential confounders and PAT output across all outcomes.

| Variable | β | SE | P-Value | Sig. |
|---|---|---|---|---|
| ***Benzodiazepine Use (n=1254)*** | | | | |
| **PAT Output** | **0.5891** | **0.2084** | **0.0047** | ** |
| PHQ9 Score | 0.6658 | 0.1028 | <0.001 | *** |
| Age | 0.2548 | 0.1737 | 0.1425 | |
| Sex | 1.5503 | 0.4019 | <0.001 | *** |
| BMI | 0.0027 | 0.1197 | 0.982 | |
| ***SSRI Use (n=1231)*** | | | | |
| **PAT Output** | **0.3215** | **0.1335** | **0.0161** | * |
| PHQ9 Score | 0.3137 | 0.0920 | 0.0006 | *** |
| Age | 0.4069 | 0.1268 | 0.0013 | ** |
| Sex | 0.5291 | 0.2452 | 0.0310 | * |
| BMI | 0.1359 | 0.0996 | 0.1725 | |
| ***Sleep Disorder History (n=1758)*** | | | | |
| **PAT Output** | **0.1647** | **0.1000** | **0.0995** | |
| PHQ9 Score | 0.4319 | 0.0677 | <0.001 | *** |
| Age | 0.2504 | 0.1015 | 0.0137 | * |
| Sex | -0.5420 | 0.1743 | 0.0019 | ** |
| BMI | 0.5949 | 0.0750 | <0.001 | *** |
| ***Sleep Abnormalities (n=1751)*** | | | | |
| **PAT Output** | **0.4408** | **0.0922** | **<0.001** | *** |
| PHQ9 Score | 0.5692 | 0.0631 | <0.001 | *** |
| Age | 0.0793 | 0.0846 | 0.3487 | |
| Sex | -0.2835 | 0.1616 | 0.0793 | |
| BMI | 0.3656 | 0.0717 | <0.001 | *** |
| ***Depression (n=1973)*** | | | | |
| **PAT Output** | **0.4124** | **0.0838** | **<0.001** | *** |
| Age | 0.2223 | 0.0778 | 0.0043 | ** |
| Sex | 0.6614 | 0.1654 | <0.001 | *** |
| BMI | 0.1729 | 0.0700 | 0.0135 | * |

*Standardized beta coefficients (β), standard errors (SE), p-values, and significance levels are shown for each outcome. Continuous variables were standardized prior to analysis. Significance: * p<0.05, ** p<0.01, *** p<0.001. n denotes the number of participants after intersecting the test set with variables shown.*

## Supplementary Table 3. Computational efficiency benchmarks for PAT and baseline models.

| Model | Params | MFLOPs | Inference (ms) |
|---|---|---|---|
| 1D CNN | 10K | 65.3 | 8.47 |
| 3D CNN | 790K | 138.7 | 7.15 |
| **PAT-S (ours)** | **287K** | **1039.5** | **21.86** |
| **PAT-M (ours)** | **1.00M** | **4038.7** | **34.64** |
| **PAT-L (ours)** | **1.99M** | **27859.7** | **62.31** |
| LSTM | 15K | 205.6 | 117.18 |
| ConvLSTM | 1.76M | 1993.3 | 158.85 |

*Inference time (ms), parameter count, and MFLOPs reported for all models on an NVIDIA T4 GPU (16 GB, Google Colab). Results shown for single-sample inference (n=1), reflecting a typical deployment scenario. PAT-S and PAT-M achieve practical inference times on consumer-grade hardware, with PAT-S completing single-sample inference in under 22 ms.*

## Supplementary Figure 1. Daily mean actigraphy attention overlay for PAT-M trained on smoothed and raw input.

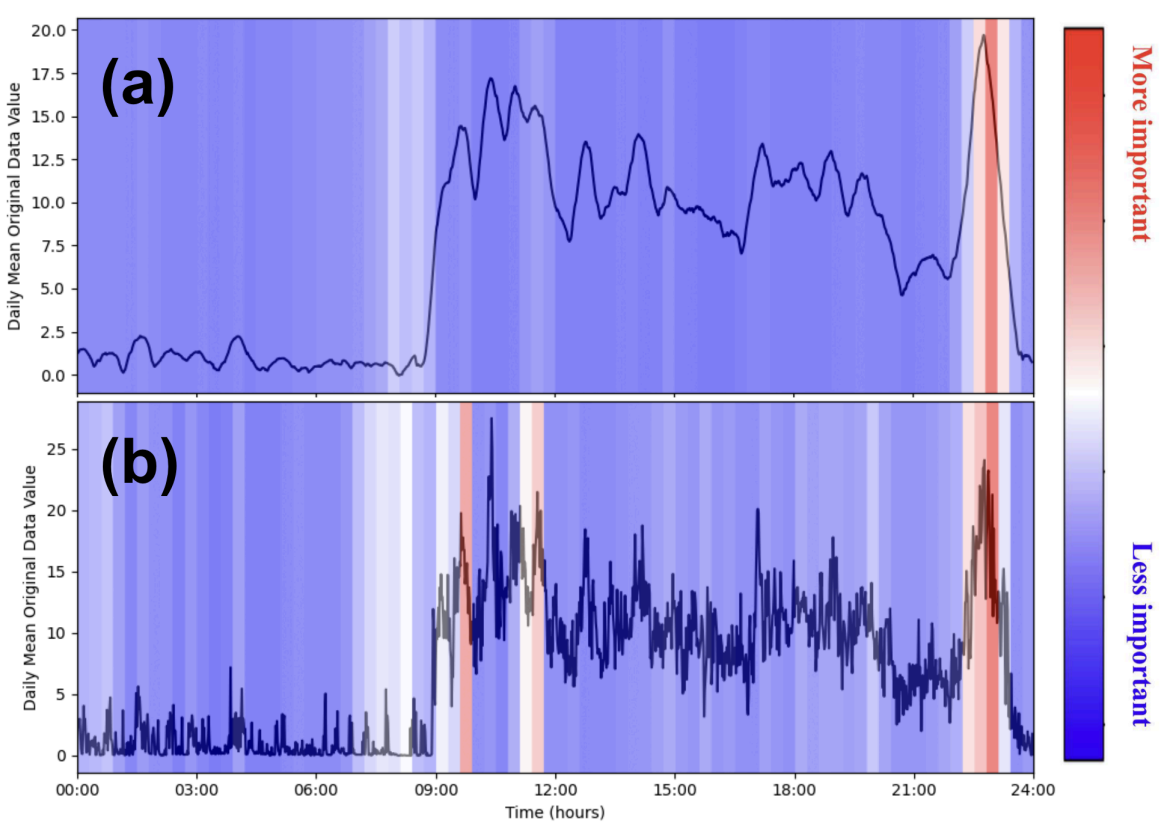

*Averaging actigraphy across the week may have made the graph appear more smoothed than it actually is. (a) Smoothed-input model. (b) Raw-input model. In each panel, the trace represents a participant's mean 24-hour activity profile averaged across the full week, with the attention overlay shown on the same time axis. This participant is taking benzodiazepines and PAT was fine-tuned to predict this outcome.*